\documentclass{article}

\everypar{\looseness=-1}

     \usepackage[final,nonatbib]{neurips_2022}

\usepackage[utf8]{inputenc} 
\usepackage[T1]{fontenc}    
\usepackage{hyperref}       
\usepackage{url}            
\usepackage{booktabs}       
\usepackage{amsfonts}       
\usepackage{nicefrac}       
\usepackage{microtype}      
\usepackage{xcolor}         
\usepackage{graphicx}       

\usepackage{amsmath}
\usepackage{amssymb}
\usepackage{mathtools}
\usepackage{amsthm}
\usepackage{blindtext} 
\usepackage{multirow}
\usepackage{algorithm}
\usepackage{algorithmic}
\usepackage{bbm}
\mathchardef\mhyphen="2D
\usepackage{authblk}
\usepackage{appendix}

\theoremstyle{plain}
\newtheorem{theorem}{Theorem}[section]
\newtheorem{proposition}[theorem]{Proposition}

\theoremstyle{definition}
\newtheorem{definition}[theorem]{Definition}

\theoremstyle{remark}

\newcommand{\X}{\mathcal{X}}
\newcommand{\Z}{\mathcal{Z}}
\newcommand{\Y}{\mathcal{Y}}
\newcommand{\T}{\mathcal{T}}
\newcommand{\Tlam}{\mathcal{T}_{\lambda}}

\newcommand{\lhat}{\hat{\lambda}}
\newcommand{\R}{\mathbb{R}}
\newcommand{\ie}{\textit{i.e.}}
\renewcommand{\P}{\mathbb{P}}
\newcommand{\E}{\mathbb{E}}
\newcommand{\ind}[1]{\mathbbm{1}\left\{#1\right\}}

\date{}

\title{Semantic uncertainty intervals for disentangled latent spaces}

\author[1]{Swami Sankaranarayanan}
\author[2]{Anastasios N.~Angelopoulos}
\author[2]{\\Stephen Bates}
\author[3]{Yaniv Romano}
\author[1]{Phillip Isola}
\affil[1]{MIT}
\affil[2]{University of California, Berkeley}
\affil[3]{Technion---Israel Institute of Technology}

\begin{document}

\maketitle

\begin{abstract}
    Meaningful uncertainty quantification in computer vision requires reasoning about semantic information---say, the hair color of the person in a photo or the location of a car on the street.
    To this end, recent breakthroughs in generative modeling allow us to represent semantic information in disentangled latent spaces, but providing uncertainties on the semantic latent variables has remained challenging.
    In this work, we provide principled uncertainty intervals that are guaranteed to contain the true semantic factors for any underlying generative model.
    The method does the following: (1) it uses quantile regression to output a heuristic uncertainty interval for each element in the latent space (2) calibrates these uncertainties such that they contain the true value of the latent for a new, unseen input.
    The endpoints of these calibrated intervals can then be propagated through the generator to produce interpretable uncertainty visualizations for each semantic factor. This technique reliably communicates semantically meaningful, principled, and instance-adaptive uncertainty in inverse problems like image super-resolution and image completion. Code and demos can be found on our \href{https://swamiviv.github.io/semantic_uncertainty_intervals/}{project page}.
\end{abstract}

\section{Introduction}\label{sec:intro}

When making decisions with visual data, such as automated vehicle navigation with blurry images, uncertainty quantification is critical.
The relevant uncertainty pertains to a low-dimensional set of semantic properties, such as the locations of objects.
However, there is a wide class of image-valued estimation problems---from super-resolution to inpainting---for which there does not currently exist a method of producing semantically meaningful uncertainties.
As an example, imagine doing uncertainty quantification for medical image reconstruction from, say, a fast but undersampled MRI scan. 
In such a setting, pixelwise intervals~\cite{gal2016dropout,oala2020interval,angelopoulos2022image} are not very useful. 
We need uncertainty on the underlying semantics---e.g., whether there is a tumor, and if so, of what size and shape.

\begin{figure}[t]
\vskip 0.2in
\begin{center}
\centerline{
\includegraphics[width=0.7\linewidth]{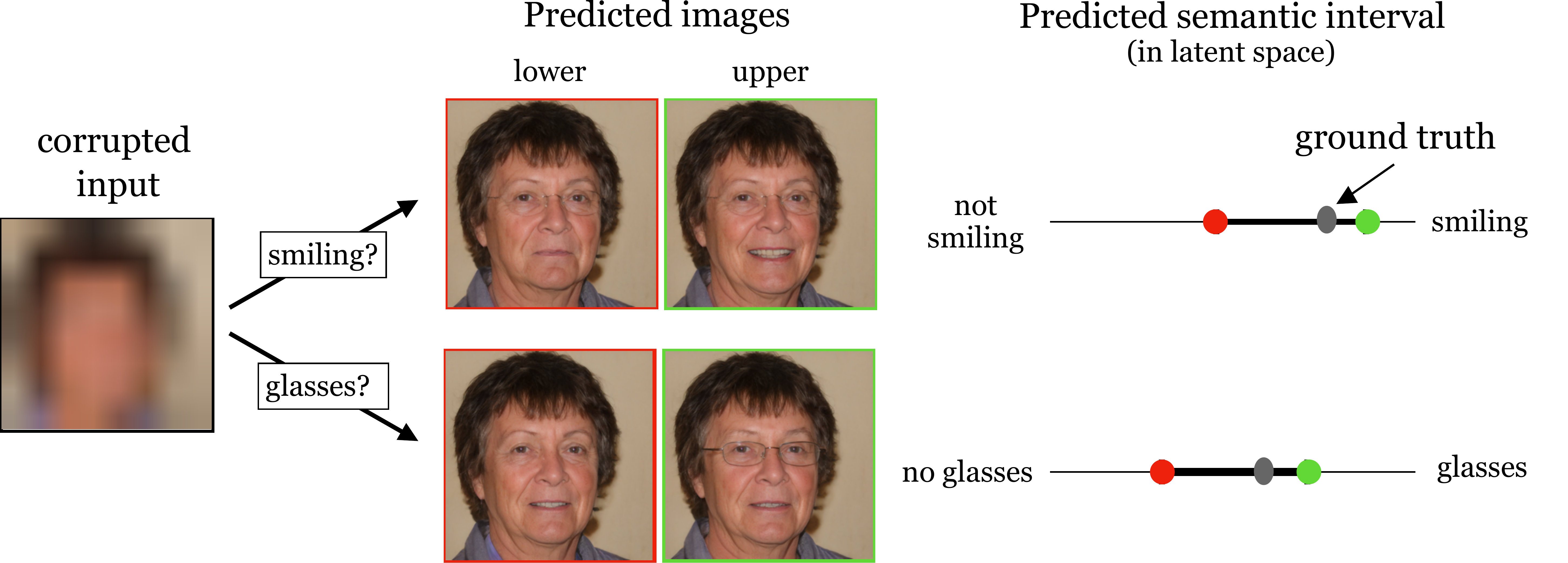}}
\caption{ \textbf{Uncertainty intervals over semantic factors} produced by our method.
We express the intervals in a disentangled latent space that allows us to factorize the uncertainty into meaningful components.}
\label{fig:teaser}
\end{center}
\end{figure}

We make progress on this problem by bringing techniques from quantile regression and distribution-free uncertainty quantification together with a disentangled latent space learned by a \emph{generative adversarial network} (GAN).
We call the coordinates of this latent space \emph{semantic factors}, as each controls one meaningful aspect of the image, like age or hair color.
We do not require these semantic factors to be statistically independent.
Our method takes a corrupted image input and predicts each semantic factor along with an uncertainty interval that is guaranteed to contain the true semantic factor.
When the model is unsure, the intervals are large, and vice-versa.
By propagating these intervals through the GAN coordinate-wise, we can visualize uncertainty directly in image-space without resorting to per-pixel intervals---see Figure~\ref{fig:teaser}.
The result of our procedure is a rich form of uncertainty quantification directly on the estimates of semantic properties of the image.

More concretely, we receive input images $X$ and then predict an \emph{uncertainty interval} for each of $D$ \emph{semantic factors} $Z_d$, $d=1,...,D$, which are the elements of a disentangled latent space.
The method involves training an \emph{encoder} (a neural network that takes images as input and produces outputs in the latent space of the GAN) on $(X,Z)$ pairs to give us three different outputs:
\begin{enumerate}
    \item \textbf{The point prediction,} $f(X)$. This is the encoder's best guess at the semantic factors $Z$.
    \item \textbf{The estimated lower conditional quantile,} $q_{\frac{\alpha}{2}}(X)$. The encoder believes that $q_{\frac{\alpha}{2}}(X)$ is a lower bound on the value of $Z$ given $X$.
    \item \textbf{The estimated upper conditional quantile,} $q_{1-\frac{\alpha}{2}}(X)$. The encoder believes that $q_{1-\frac{\alpha}{2}}(X)$ is an upper bound on the value of $Z$ given $X$.
\end{enumerate}

Once the above encoder is trained, as described in Section~\ref{subsec:training}, we use it to form an uncertainty interval for each semantic factor.
However, for the $d$th element of the latent code, the naive interval $(q_{\frac{\alpha}{2}}(X)_d,q_{1-\frac{\alpha}{2}}(X)_d)$ is not guaranteed to contain the ground truth value in finite samples.
We propose to perform a calibration procedure to fix this problem, yielding the sets
\begin{equation}
    \T(X)_d = \left[q^{\rm cal}_{\frac{\alpha}{2}}(X)_d, q^{\rm cal}_{1-\frac{\alpha}{2}}(X)_d\right],
\end{equation}
where $q^{\rm cal}$ is a calibrated version of $q$ constructed using the tools in Section~\ref{subsec:calibration}.
Once we have done so, the intervals will contain $\alpha$ fraction of the true latent codes with high probability.
In other words, for any user-chosen levels $\alpha$ and $\delta$, we can output intervals that with probability $1-\delta$ satisfy
\begin{equation}
    \label{eq:interval-form}
    \E\left[\frac{1}{D}\Big|\big\{d : Z_{d} \in \T(X)_d\big\}\Big|\right] \ge 1 - \alpha,
\end{equation}
for a new test point $(X,Z)$, regardless of the distribution of the data, the encoder used, and the number of data points used in the calibration procedure.
This guarantee, described more carefully in Definition~\ref{def:rcps}, says that the intervals cover $1-\alpha$ fraction of the semantic factors unless our calibration data is not representative of our test data (which only happens with a probability $\delta$ which goes to $0$ as the number of calibration data points grows).
We visualize each of the $d \in \{1,...,D\}$ intervals in latent space by propagating the $d$th lower and upper endpoints through the generator with all other entries in the latent fixed to the point prediction (see Section~\ref{subsec:visualization} for a formal explanation). 

\subsection{Central Contribution}

To our knowledge, this is the first algorithm for uncertainty intervals on a learned semantic representation with formal statistical guarantees.
By propagating these intervals through the generator, we are able to visualize uncertainty in an intuitive new way that directly encodes semantic meaning.
This is an important step towards interpretable uncertainties in general image-valued estimation problems.
However, the reader should note that our technique requires access to a disentangled latent space, such as that of a StyleGAN, and provides no guarantees about the degree of disentanglement.

\section{Method}
\label{sec:method}
\subsection{Notation and goal}


Our data consist of pairs $(X,Z)$---the corrupted image $X$ in $\X=[0,1]^{H\times W}$, and the latent code $Z \in \Z$, where $\Z = \R^D$.
As mentioned in the introduction, we think of $Z$ as a disentangled representation with $d$ \emph{semantic factors}---\ie  factors of variation corresponding to interpretable features in an image, such as hair color and expression.
For simplicity, assume each of the $d$ dimensions controls a single semantic factor; in practice, we ignore those that do not.

In our sampling model, $X$ is generated from $Z$ by composing two functions.
The first function is a fixed generator $G : \Z \to \Y$, where $\Y=[0,1]^{H \times W}$, which takes the latent vector $Z$ and produces the ground truth image $Y \in \Y$ (for ease of notation, we assume $X$ and $Y$ have the same shape).
The second function is a corruption model, $F : \Y \to \X$, which degrades the ground truth image $Y$ to produce the corrupted image $X$, for example by randomly masking out part of the image.
To summarize our data-generating process, we have
\begin{equation}
    Y = G(Z)\text{ and } X = F(Y);
\end{equation}

\textbf{Goal \#1.} Our first goal is to train an encoder $E$ to recover $Z$ from $X$---in other words, to invert the mapping $F\circ G$---with a heuristic notion of uncertainty.
The encoder's point prediction will be a function $f : \X \to \Z$.
The uncertainty will be parameterized by two functions, $q_{1-\frac{\alpha}{2}} : \X \to \Z$ and $q_{\frac{\alpha}{2}} : \X \to \Z$, denoting our estimates of the $1-\frac{\alpha}{2}$ and $\frac{\alpha}{2}$ conditional quantiles, respectively.
These conditional quantiles are potentially bad estimates; they do not natively possess the statistical guarantee we desire.

\textbf{Goal \#2.} Having trained the encoder and the conditional quantile estimates, we will output uncertainty intervals in the disentangled latent space.
Each dimension will get its own interval, which has the form in~\eqref{eq:interval-form}.
Ultimately, our uncertainty intervals
will come with a statistical risk control guarantee, like in (2), that is distribution-free---i.e., valid irrespective of the model or data distribution.
The risk function controlled is the false negative rate \emph{in the latent space}.
In other words, we bound the fraction of semantic factors not covered by their calibrated intervals above by $\alpha$ (say, $\alpha=10\%$).
\begin{definition}[Risk-Controlling Prediction Set (RCPS)]
    \label{def:rcps}
    A set-valued function $\T : \X \to 2^\Z$ is an $(\alpha,\delta)$ risk-controlling prediction set if 
    \begin{equation}
        \label{eq:rcps-guarantee}
        \P\Big(\E\big[L(\T(X),Z)\big] > \alpha \Big) \leq \delta, 
    \end{equation}
    where
    \begin{equation}        L(\T(X),Y) = 1-\frac{\Big|\big\{d : Z_d \in \T(X)_d\big\}\Big|}{HW}.
    \end{equation}
\end{definition}
The reader should note here that the function $\T$ depends on the calibration data.
The outer probability in~\eqref{eq:rcps-guarantee} is over the randomness in this calibration procedure; the inner expectation is over the new test point, $(X,Y)$.
The reader should note that in our setting, because we can generate infinite data from the GAN, it is always possible to take $\delta$ arbitrarily close to $0$, effectively making $\T$ nonrandom. 
In the two following subsections, we address each of our goals separately.

\subsection{Goal \#1: Training the encoder for quantile regression}
\label{subsec:training}

\begin{figure}[t]
\begin{center}
\centerline{
\includegraphics[width=0.45\columnwidth]{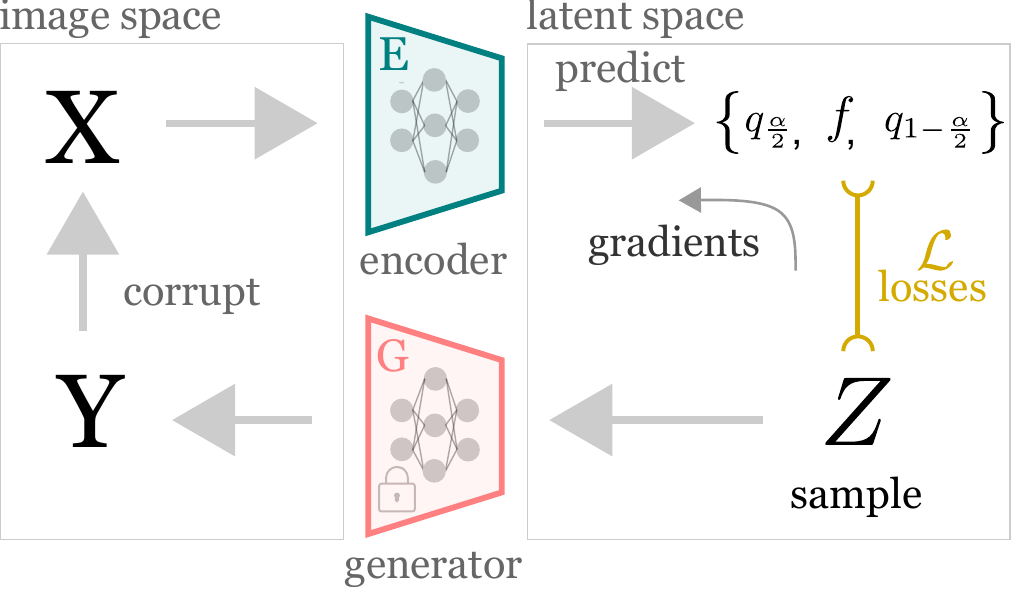}
}
\caption{\textbf{Our training pipeline}, visualized above, shows our training process for the prediction $\hat{f}$, lower quantile $q_{\frac{\alpha}{2}}$ and the upper quantile $q_{1-\frac{\alpha}{2}}$.}
\label{fig:training}
\end{center}
\end{figure}

Our job in this subsection is to learn the three functions $f$, $q_{\frac{\alpha}{2}}$, and $q_{1-\frac{\alpha}{2}}$.
We do so by training a neural network with three different loss functions, one for each of three linear heads on top of the same feature extractor---see Figure~\ref{fig:training} for the training protocol.


\textbf{Loss function for point prediction.} We supervise the point prediction with three loss functions.
The first is an $L_1$ loss directly in the latent space, and encourages $f(X)$ to be close to $Z$.
\begin{equation}
    \label{eq:l1-loss}
    \mathcal{L}_1\big(f(x),z\big) = \big|\big|f(x)-z\big|\big|_1
\end{equation}
The second loss term imposes a \emph{domain specific prior} on the generated image $G(f(X))$. For our experiments on faces, we use an identity loss which encourages $G(f(X))$ to contain attributes that have the same ``identity'' --- or semantic content --- as $Y$.
We calculate the identity loss using a pretrained embedding function, $\mathrm{ID} : \Y \to \R^{d'}$ for some $d'$, which projects $Y$ to an embedding space where images with different identities land far away from one another.
Finally, we calculate our loss using cosine similarity,
\begin{equation}
    \label{eq:id-loss}
    \mathcal{L}_{\rm ID}\big(x, y\big) = \frac{ \langle \mathrm{ID} (x), \mathrm{ID} (y) \rangle }{||\mathrm{ID} (x)|| \cdot ||\mathrm{ID} (y)||},
\end{equation}

The third is a \emph{perceptual loss} on the generated image $G(f(X))$ that imposes similarity between $G(f(X))$ and $X$ in the embedding space of a large pretrained image classification model. We impose perceptual similarity by computing the LPIPS loss~\cite{lpips} which has been shown to preserve image quality~\cite{lpips_proof}. The perceptual loss is calculated using a pretrained feature extractor in a similar manner to Eq~\ref{eq:id-loss}. Following standard practice, the feature extractor used in this case is a VGG network, pretrained on Imagenet.

Finally, we combine the loss functions to form the loss function for the prediction $f$, 
\begin{equation}
    \label{eq:full-loss}
    \mathcal{L}_{\rm pred}\big(f(x),z\big) = \mathcal{L}_1\big(f(x),z\big) + c_1\mathcal{L}_{\rm ID}\big(G(f(x)),G(z)\big) + c_2\mathcal{L}_{\rm LPIPS}\big(G(f(x)),G(z)\big),
\end{equation}
where $c_1$ and $c_2$ are hyper-parameters whose values are chosen based on a held out set.

\textbf{Loss function for quantile regression.}  Quantile regression~\cite{koenker1978regression,chaudhuri1991global,koenker2001quantile,koenker2005quantile,koenker2011additive,koenker2017quantile,koenker2018handbook} is a statistical method for estimating the conditional quantiles of a distribution.
The key idea of quantile regression is to supervise the regressor using a \emph{quantile loss},
\begin{equation}
    \label{eq:quantile-loss}
    \begin{aligned}
        \mathcal{L}^{\beta}_{\rm q}(q_{\beta}(x),z) = \big(z-q_{\beta}(x)\big)\beta
        \ind{z > q_{\beta}(x)} + \big(q_{\beta}(x) - z\big)(1-\beta)
        \ind{z \leq q_{\beta}(x)}.
    \end{aligned}
\end{equation}
The minimizer of the quantile risk is the true $\beta$ conditional quantile of $Z|X$.
We supervise our conditional quantile estimates $q_{\frac{\alpha}{2}}$ and $q_{1-\frac{\alpha}{2}}$ with two separate instances of the quantile loss, $\mathcal{L}^{\frac{\alpha}{2}}_{\rm q}$ and $\mathcal{L}^{1-\frac{\alpha}{2}}_{\rm q}$ respectively.

This concludes our explanation of the model training procedure, summarized in Algorithm~\ref{alg:gqan-encoder-training}.
Experimental details, such as the particular model architecture we use, are available in Section~\ref{sec:experiments}. 
\begin{algorithm}[H]
   \caption{Quantile GAN encoder training}
   \label{alg:gqan-encoder-training}
\begin{algorithmic}
   \STATE {\bfseries Input:} training dataset $\mathcal{D}$, risk level $\alpha$, number of epochs $E$, fixed generator $G$
   \STATE {\bfseries Output:} trained functions $f,q_{\frac{\alpha}{2}},$ and $q_{1-\frac{\alpha}{2}}$.
   \STATE $f,q_{\frac{\alpha}{2}},$ and $q_{1-\frac{\alpha}{2}} \gets$ random model initialization
   \FOR{$e=1$ {\bfseries to} $E$}
      \STATE \texttt{loss} $\gets 0$
      \FOR{$(X^{\rm train},Z^{\rm train})$ {\bfseries in} $\mathcal{D}$}
          \STATE \texttt{L1} $\gets \mathcal{L}_{\rm pred}\big(f(X^{\rm train}\big),Z^{\rm train})$ 
          
          \STATE \texttt{L2} $\gets \mathcal{L}^{\frac{\alpha}{2}}_{\rm q}\big(q_{\frac{\alpha}{2}}(X^{\rm train}\big),Z^{\rm train})$
          
          \STATE \texttt{L3} $\gets \mathcal{L}^{1-\frac{\alpha}{2}}_{\rm q}\big(q_{1-\frac{\alpha}{2}}(X^{\rm train}\big),Z^{\rm train})$
          
          \STATE \texttt{loss += L1 + L2 + L3}
      \ENDFOR
      \STATE \texttt{loss.backward()}
   \ENDFOR

\end{algorithmic}
\end{algorithm}

\subsection{Goal \#2: Calibration}
\label{subsec:calibration}

Having trained the model, we now calibrate it to achieve the statistical guarantee in Definition~\ref{def:rcps} using a set of calibration data $\big\{(X_i,Z_i)\big\}_{i=1}^n$ generated from the model and the upper-confidence bound procedure from~\cite{bates2021distribution}.
The output of the procedure will be the function $\T$ from~\eqref{eq:interval-form}; specifically, we will learn the calibrated conditional quantiles $q^{\rm cal}_{\frac{\alpha}{2}}$ and $q^{\rm cal}_{1-\frac{\alpha}{2}}$.


Our procedure will calibrate the conditional quantiles by rescaling their size multiplicatively.
We will ultimately choose a multiplicative factor $\lhat$ that gives us the desired guarantee.
Towards that end, we index a family of uncertainty intervals scaled by a free parameter $\lambda$ for each semantic factor,
\begin{equation}
    \begin{aligned}
        \T_{\lambda}(X)_d = \bigg[ f(X)_d - \lambda \big(f(X)_d-q_{\frac{\alpha}{2}}(X)_d\big)_+, \; \; \;
        f(X)_d + \lambda \big(q_{1-\frac{\alpha}{2}}(X)_d-f(X)_d\big)_+ \bigg].
    \end{aligned}
    \label{eq:uncertainty_interval}
\end{equation}
When $\lambda$ grows, the interval $\Tlam(X)_d$ also grows, and thus, the loss function $L(\Tlam(X),Y)$ shrinks.
Therefore, by taking $\lambda$ large enough, we can always ensure the loss is zero.
The challenge ahead is to pick $\lhat$ to be the smallest value such that $\T_{\lhat}$ is an RCPS as in~\eqref{eq:rcps-guarantee}.

The algorithm for selecting $\lhat$ involves forming an upper confidence bound (UCB) for the risk, then picking the smallest value of $\lambda$ such that the upper confidence bound falls below $\alpha$.
We give Hoeffding's UCB below, although we use the stronger Hoeffding-Bentkus bound from~\cite{bates2021distribution} in practice:
\begin{equation}
    \label{eq:hoeffding}
    \hat{R}^+(\lambda) = \frac{1}{n}\sum\limits_{i=1}^n L\left( \Tlam(X_i), Y_i \right) + \sqrt{\frac{1}{2n}\log \frac{1}{\delta}}.
\end{equation}
Note that in our setting, we can always generate enough samples to drive the last term to $0$ for any $\delta$; however, if we only had a finite sample from some population, this would not be the case.
We can then select $\lhat$ by scanning from large to small values, $\lhat = \min\left\{\lambda : \hat{R}^+(\lambda') \leq \alpha, \;\; \forall \alpha' \geq \alpha \right\}$, or running binary search.
\begin{proposition}[$\T_{\lhat}$ is an RCPS~\cite{bates2021distribution}]
    \label{prop:rcps-guarantee}
    With $\lhat$ selected as above, $\T_{\lhat}$ satisfies Definition~\ref{def:rcps}.
\end{proposition}
For the proof of this fact, along with a discussion the tighter confidence bounds used in our experiments and extensions to the underlying theory, see~\cite{bates2021distribution} and~\cite{angelopoulos2021learn}.


Having proven that $\T_{\lhat}$ is an RCPS, we can simply set $\T(X) = \T_{\lhat}(X)$ in~\eqref{eq:interval-form}; in other words, we set $q^{\rm cal}_{\frac{\alpha}{2}}(X)_d = f(X)_d - \lhat \big(f(X)_d-q_{\frac{\alpha}{2}}(X)_d\big)_+$ and $q^{\rm cal}_{1-\frac{\alpha}{2}}(X)_d = f(X)_d + \lhat \big(q_{1-\frac{\alpha}{2}}(X)_d-f(X)_d\big)_+$.

\subsubsection*{Visualizing uncertainty intervals in image space}
\label{subsec:visualization}
We briefly describe our method for visualizing latent-space uncertainty intervals.
In order to see the effect of a single semantic factor, we set it to either the lower or upper quantile and hold the other factors fixed to the point prediction. More specifically, for a particular dimension $d \in \{1,...,D\}$, define the following vector:
\begin{equation}
\begin{aligned}
     \hat{Z}^d_{k} =  \big(f(X)_1,...,f(X)_{d-1},\;\;\; q^{cal}_{k}(X)_d,\;\;\;f(X)_{d+1},...,f(X)_{D}\big).
\end{aligned}
\end{equation}
We visualize the lower and upper quantiles in image space as $G(\hat{Z}^{d}_{\frac{\alpha}{2}})$ and $G(\hat{Z}^{d}_{1-\frac{\alpha}{2}})$ respectively.
Since each semantic factor corresponds to an attribute, visualizing the lower and upper quantiles per-factor gives interpretable meaning to the latent-space intervals.
For example, in Figure~\ref{fig:teaser}, the images of the child smiling give a range of possible expressions the model believes are consistent with the underlying image.


\section{Experiments}\label{sec:experiments}

\subsection{Dataset descriptions}

\textbf{FFHQ} We use the StyleGAN framework pretrained using the Flickr-Faces-HQ (FFHQ) dataset~\cite{stylegan1_2019}. FFHQ is a publicly available dataset consisting of 70,000 high-quality images at $1024 \times 1024$ resolution. The data used to train the quantile encoder and for the experiments in this section is sampled from the generator pretrained on FFHQ.

\textbf{CelebA-HQ}  We use the CelebA-HQ dataset ~\cite{karras2018celeba-hq} to demonstrate the effectiveness of our approach on real world data. The dataset  contains 30,000 high-quality images of celebrity faces. Since the real dataset is used only for evaluation, we use images from standard test split.

\textbf{CLEVR.} In order to have a controlled setup where we can easily identify disentangled factors of variation, we generate synthetic images of objects based on the CLEVR dataset ~\cite{johnson2017clevr}. This dataset provides a programmatic way of generating synthetic data by explicitly varying specific semantic factors. We create a synthetic dataset by varying $\{color,shape\}$ and fixing the other factors such as lighting, material and camera jitter.
\vspace{-0.35cm}

\subsection{Experimental setup}

\textbf{Model architectures}
In all our experiments, we use the StyleGAN2 ~\cite{karras2020stylegan2} framework for the generator architecture $G$. For the experiments involving faces, we use the pretrained model available from the official repository. For the CLEVR-2D experiments, we train a simpler variant of StyleGAN2 generator from scratch. For the quantile regression, we use a standard architecture: the encoder network consists of a ResNet-50 backbone~\cite{he2016deep} with the final layer branching into the point prediction and conditional quantiles. The branching module is a standard combination of convolution and activation blocks followed by a fully connected layer of the expected output dimension; we call these the \emph{heads} of the model. Specific details of the model architecture are provided in the supplementary material.

\textbf{Model training}
We start by pretraining the generative model or acquiring an off-the-shelf pretrained generative model for the task at hand. 
In generative models such as StyleGAN, the style space that offers fine grained control over image attributes, is very high dimensional. From this high dimensional space, we extract only the disentangled dimensions following previous work on style space analysis ~\cite{elad2021psp}. In order to better focus the encoder's capacity only on the disentangled dimensions, we mask out the irrelevant dimensions for applying the quantile loss. However, the pointwise loss in ~\eqref{eq:l1-loss} is applied to the full style vector to ensure that the pointwise prediction is able to match the true latents accurately, while the quantile heads focus on learning variablity only in the disentangled dimensions.

During the encoder training ~\ref{subsec:training}, the generative model $G$ is held frozen and only the parameters of the encoder $E$ are updated. The point prediction and conditional quantile heads are trained jointly with the Ranger optimizer ~\cite{elad2021psp} and a flat learning rate of 0.001 for all our experiments. The hyperparameter weights (Eq~\ref{eq:full-loss}) are set to $c_1=c_2=10.0$.

For the image super-resolution training, we augment the input dataset by using different levels of downsampled inputs, \ie, we take the raw input and apply a random downsampling factor from $\{1, 4, 8, 16, 32\}$ and resize it to the original dimensions. For the image inpainting task, we vary the difficulty by choosing a random threshold to create the mask -- lower thresholds implies fewer pixels are masked and vice-versa. The mask is concatenated to the image resulting in a $C+1$-channel input to the encoder, where $C$ is the number of image channels. The detailed description of the mask generation procedure can be found in the supplementary material.

\textbf{Calibration and Evaluation}
For both the synthetic object experiments with CLEVR and face experiments with FFHQ, we train the quantile encoder on data points sampled from the latent space of the pretrained generative model. This ensures that we have access to the \textit{true} latents that resulted in each image. We generate 100k samples per model and generate a random 80-10-10 split for training, calibration and validation.
\vspace{-0.3cm}

\begin{figure}[t]
\includegraphics[width=0.48\textwidth]{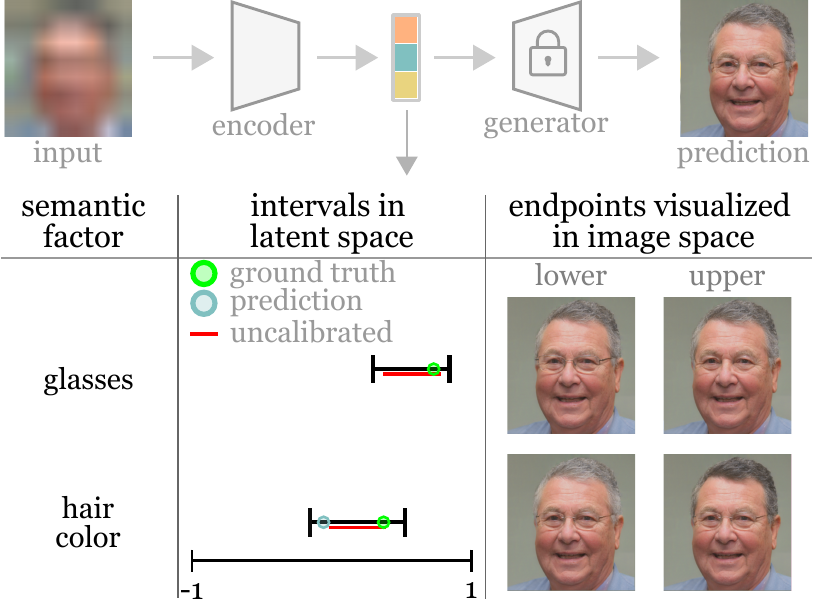}
\hfill
\includegraphics[width=0.48\textwidth]{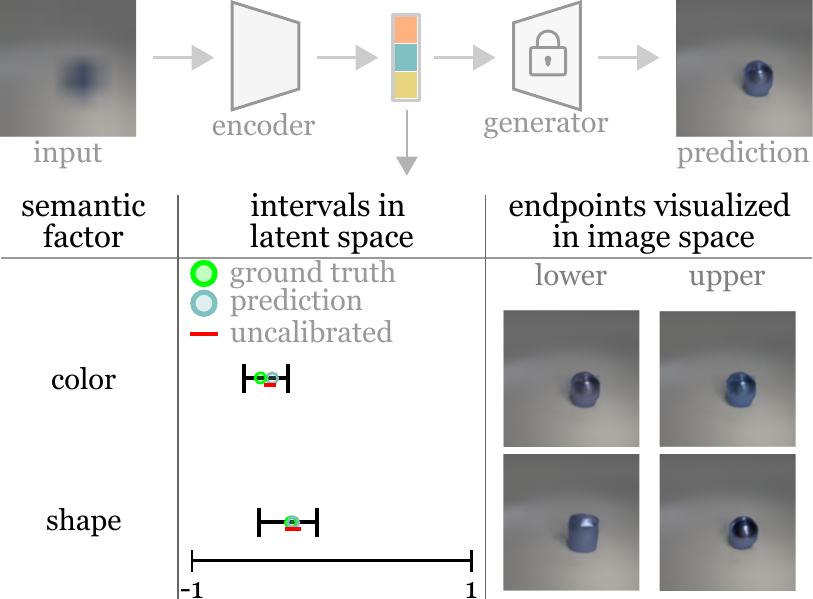}
\caption{\textbf{Semantically meaningful uncertainty intervals} produced by our method on example images sampled from the generator trained on the FFHQ dataset (left) and CLEVR dataset (right). The corrupt image is provided as input to the encoder which outputs a pointwise prediction and quantile predictions for each style dimension. 
We plot the calibrated and uncalibrated intervals as well as their visualizations in image-space. [Best viewed in color, zoom in for detail.]}
\label{fig:ffhq_clevr_predictions}
\vspace{-0.79cm}
\end{figure}

\subsection{Findings}
In the following experiments, we explore different properties of our intervals. The problem types include image super-resolution and image inpainting. The risk level $\alpha$ and the user-specified error threshold $\delta$ are fixed to 0.1, unless specified otherwise.  
\vspace{-0.25cm}

\subsubsection{Producing semantic uncertainties}\label{subsec:sem_unc}

\textbf{Goal.}  We qualitatively verify that the proposed approach outlined in Section~\ref{sec:method} produces visually meaningful uncertainties.

\textbf{Description.} We train a quantile encoder with Algorithm~\ref{alg:gqan-encoder-training}.
Then we generate $n=5000$ images by sampling latents and propagating them through the encoder-generator combination. We use these $n$ images as calibration data for the procedure in~\ref{subsec:calibration}.
Finally, we randomly sample a new test point, pass it through the calibrated encoder, and form the uncertainty intervals in image space as in Section~\ref{subsec:visualization}.

\textbf{Results.} The results are illustrated in Figure~\ref{fig:ffhq_clevr_predictions} on images sampled from the generator trained on FFHQ (left) and CLEVR (right).  In case of the FFHQ image, the person is wearing glasses in the lower quantile image and not in the upper; hence, the model is not certain that the person is wearing glasses.
The model also expresses some uncertainty about the amount of gray versus brown hair.
This outcome was predictable from the input image, where the fact that the person is wearing glasses is not obvious, and there is some hair color ambiguity. The results on the CLEVR dataset are analogous. 
The lower and upper quantile images yield similar colors, which is predictable from the blurry input.
The model predicts that both a cylinder and sphere would be consistent with this blurry input.
The calibrated quantiles cover the ground truth color value, while the uncalibrated ones do not.


\subsubsection{Experiments with real data}
\begin{figure}[H]
\includegraphics[width=0.54\textwidth]{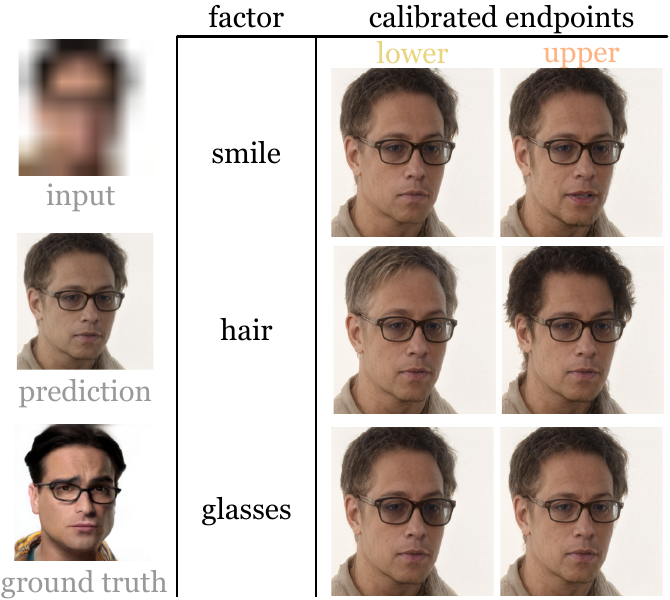} \hfill
\includegraphics[width=0.4\textwidth]{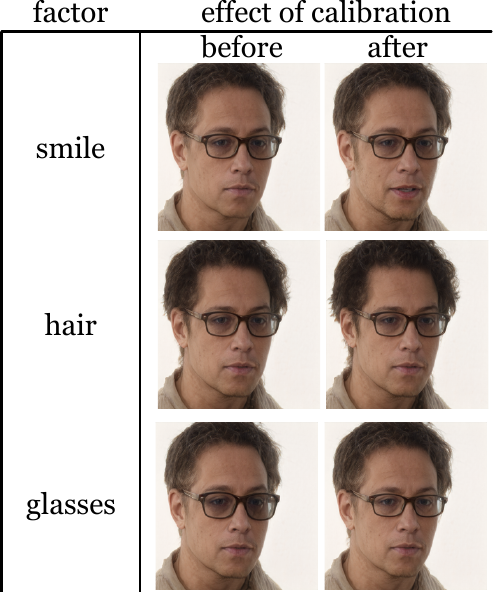}
\caption{\textbf{Uncertainty intervals for real data} We demonstrate the outputs of our approach for a real image sampled from the CelebA-HQ faces dataset. \textbf{[Left]} The input is corrupted by downsampling the true image by a factor of 32. 
Though the prediction does not match the ground truth, the uncertainty captured by the quantiles remains meaningful. 
\textbf{[Right]} For the same input, we show the effect of our calibration procedure by visualizing the edges of the quantile intervals before and after calibration. Observe that in some cases, such as \textit{smile} and \textit{glasses}, there are clear visible changes due to the calibration approach. In other cases such as \textit{hair}, the changes are not visually explicit.  
More such examples can be found in the supplementary material. [Best viewed in color. Zoom in for detail]}
\label{fig:real_data}
\end{figure}
\textbf{Goal.} We calibrate our encoder with a combination of fake data sampled from the pretrained StyleGAN model and real data sampled from the CelebA-HQ face dataset for the image resolution task. The objective is to test to what extent we can extract meaningful uncertainty intervals for real data without explicitly training with them.

\textbf{Description} The difficulty with calibrating using real data is that the targets i.e the true style vectors that generated the real images are not available to us. To circumvent that difficulty, we back-project the real images into the latent space of the GAN using existing approaches to GAN inversion~\cite{karras2020stylegan2, im2stylegan}. Even though the mapping from the GAN latent space to the output is not one-one, the back-projection optimization procedure is able to visually match the appearance of the real images convincingly. This procedure provides us with the estimated "true" style vectors for each image in the real calibration set. We combine this with the generated data and perform the RCPS calibration procedure as described in Section~\ref{subsec:calibration}. Note that, our encoder was only trained on generated data i.e. the same model as used in Section~\ref{subsec:sem_unc}, we use the real data only for the calibration procedure.

\textbf{Qualitative results} The predicted quantiles and the effect of the calibration procedure for a real test image drawn from the CelebA-HQ dataset is visualized in Figure~\ref{fig:real_data}. The pointwise prediction of the encoder does not match the identity of the true image due to the high level of input corruption. However, using our approach, we are able to visualize the predicted uncertainty intervals in a meaningful manner. The visual effect of our calibration procedure is demonstrated on the right. Notice that some attributes such as \textit{smile} and \textit{glasses} have a visible change but others do not. The change is due to the fact that the calibration procedure adjusts the intervals in such a way to satisfy the required coverage guarantee.

\subsubsection{Exploratory results with purposeful corruptions}
\begin{figure}[H]
\includegraphics[width=0.48\textwidth]{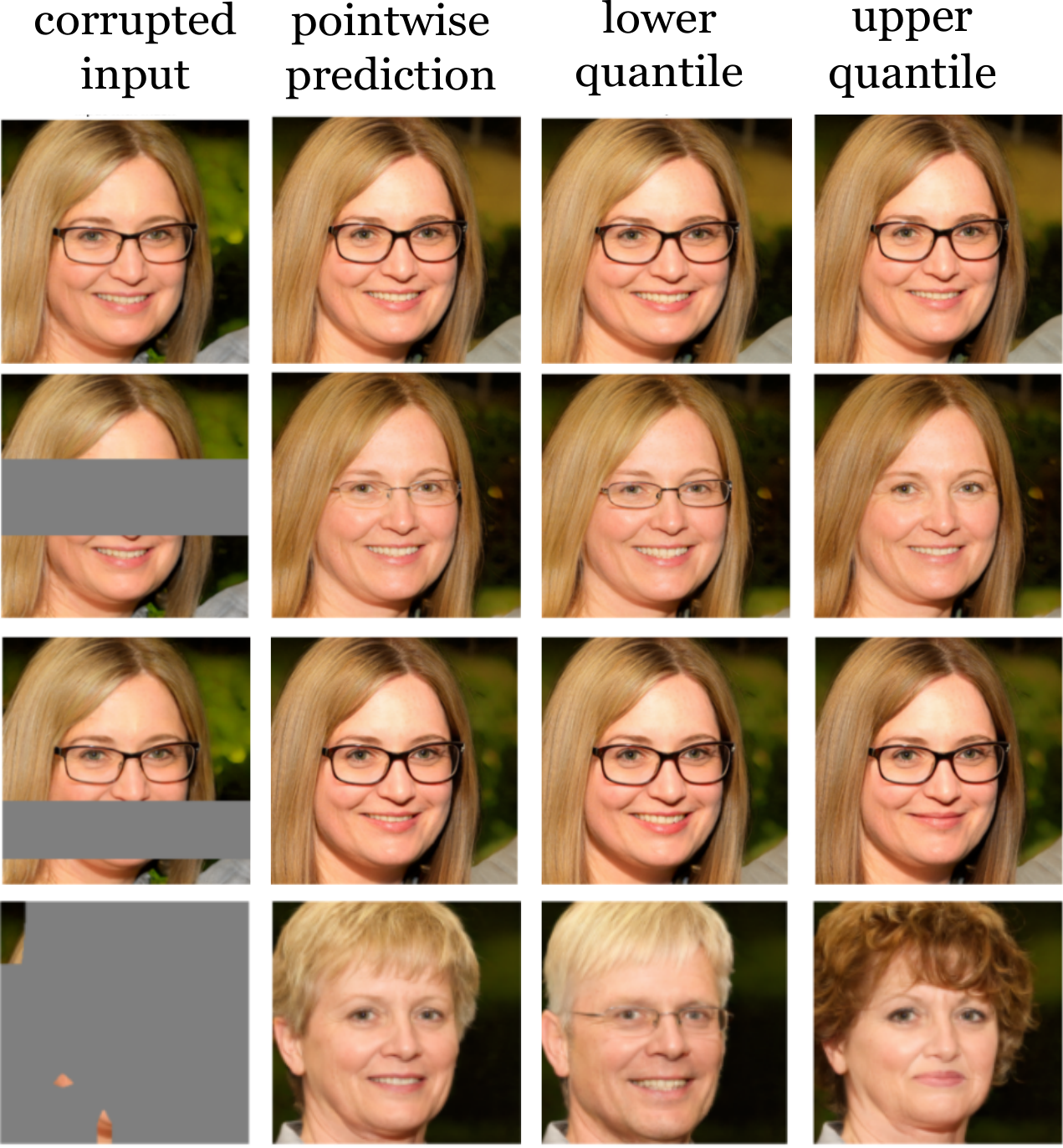} \hfill
\includegraphics[width=0.48\textwidth]{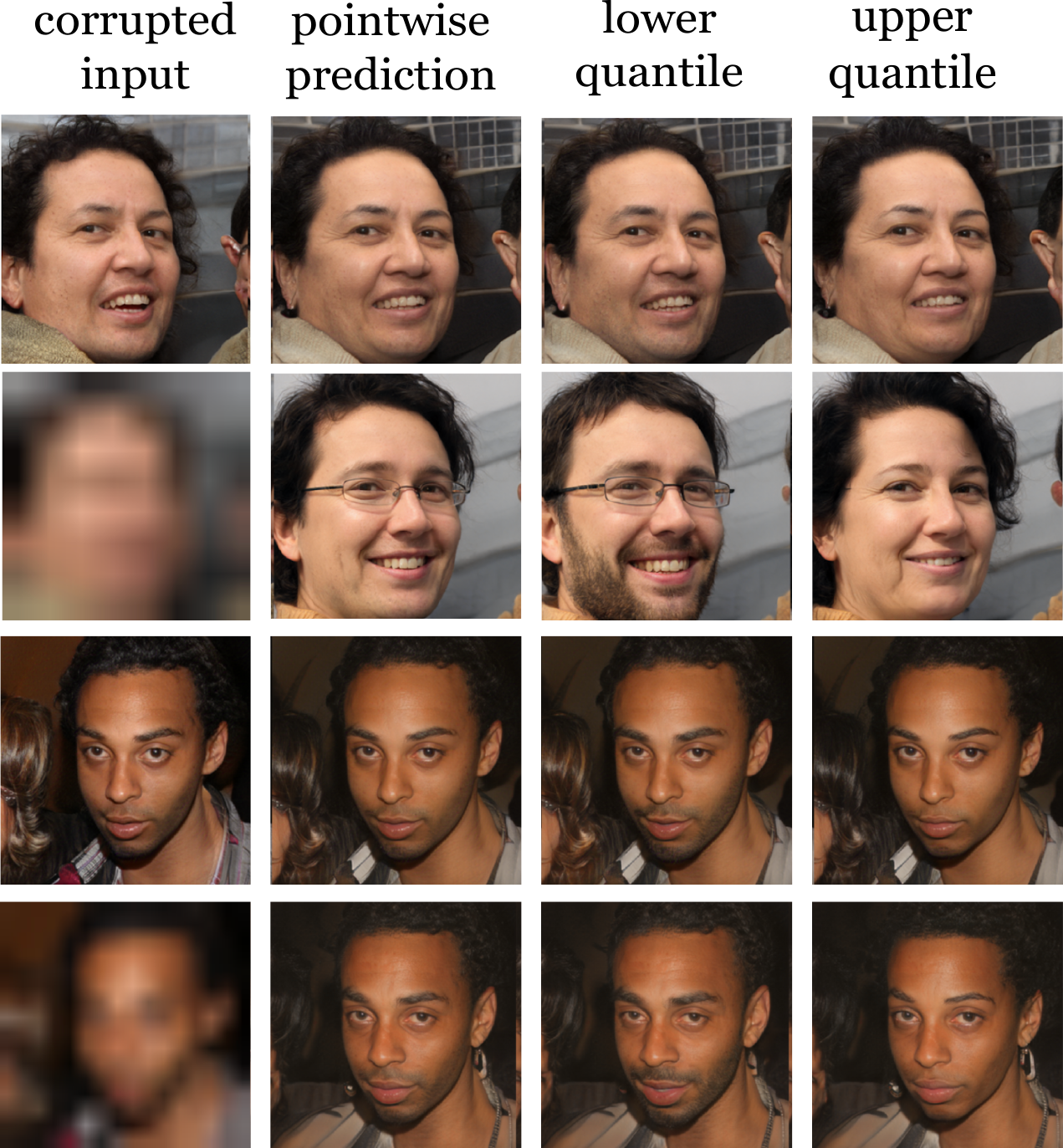}
\caption{\textbf{Visualizing adaptivity} \textbf{[Left]} A random mask is applied to the same input image in each row. When there is no mask (1st row), the lower and upper quantiles are extremely close to the pointwise prediction. As we increase the regions that are being masked, the intervals predicted by the quantile encoder expand, as indicated by the variability on the lower and upper quantile predictions. \textbf{[Right]} We show the results of the encoder on two sets of images. The corruption intensity is varied across each set, the input image in the top row is not corrupted while the input in the bottom row is under-sampled by 16x. In both case, we can observe that the most diverse prediction is in the bottom row where the input is corrupted the most. [Best viewed in color. Zoom in for detail]}
\label{fig:adaptivity}
\vskip -0.2in
\end{figure}
\textbf{Goal.} 
We probe our procedure to see if it will have the expected qualitative behavior.

\textbf{Description of experiment.}
We sampled images from the held out set of the FFHQ pretrained GAN and applied purposeful corruptions to check if the resulting quantile estimates had semantic meaning. Both image resolution and image masking were used as corruption models. We qualitatively analyze the results by visualizing the predictions. We include a quantitative measurement of variability estimated using image based metrics in the supplementary material. 

\textbf{Qualitative results.} Figure~\ref{fig:adaptivity} shows the results of this experiment for Image inpainting (left) and super-resolution (right). In the inpainting case, when nothing is masked, the quantiles are roughly identical. When the eyes are masked, the quantiles indicate the model does not know if the person was wearing glasses.  
When the mouth is masked, the model expresses uncertainty as to whether the woman is showing her teeth in the smile. Finally, when almost everything is masked, the quantile images are very different, representing individuals with entirely different identities. Similar behavior can be observed in the super-resolution case. The results are shown for two separate inputs; in both cases, the input in the top row is uncorrupted while in the bottom row, it is undersampled by 16x. The model is able to predict almost perfectly in the absence of corruption. The quantile predictions are extremely close as well. In the presence of corruption, both the pointwise prediction is off (as expected) and the quantile edges display much higher variability including in attributes such as hair shape, glasses, facial hair and perceived gender. The results from both these experiments point to an expected qualitative behavior of our proposed approach: the model exhibits more uncertainty with increased information loss at the input.
\vspace{-2mm}

\section{Interval sizes as a function of problem difficulty}
\textbf{Goal.} 
We seek to construct intervals that adapt to the uncertainty of the input, \ie, result in lower values for easier inputs and higher values for harder inputs. In this experiment we verify quantitatively how informative our intervals are as a function of increasing input corruption.
\vspace{-2mm}
\begin{figure}[H]
\begin{center}
\begin{tabular}{cc}
  \includegraphics[width=0.45\textwidth]{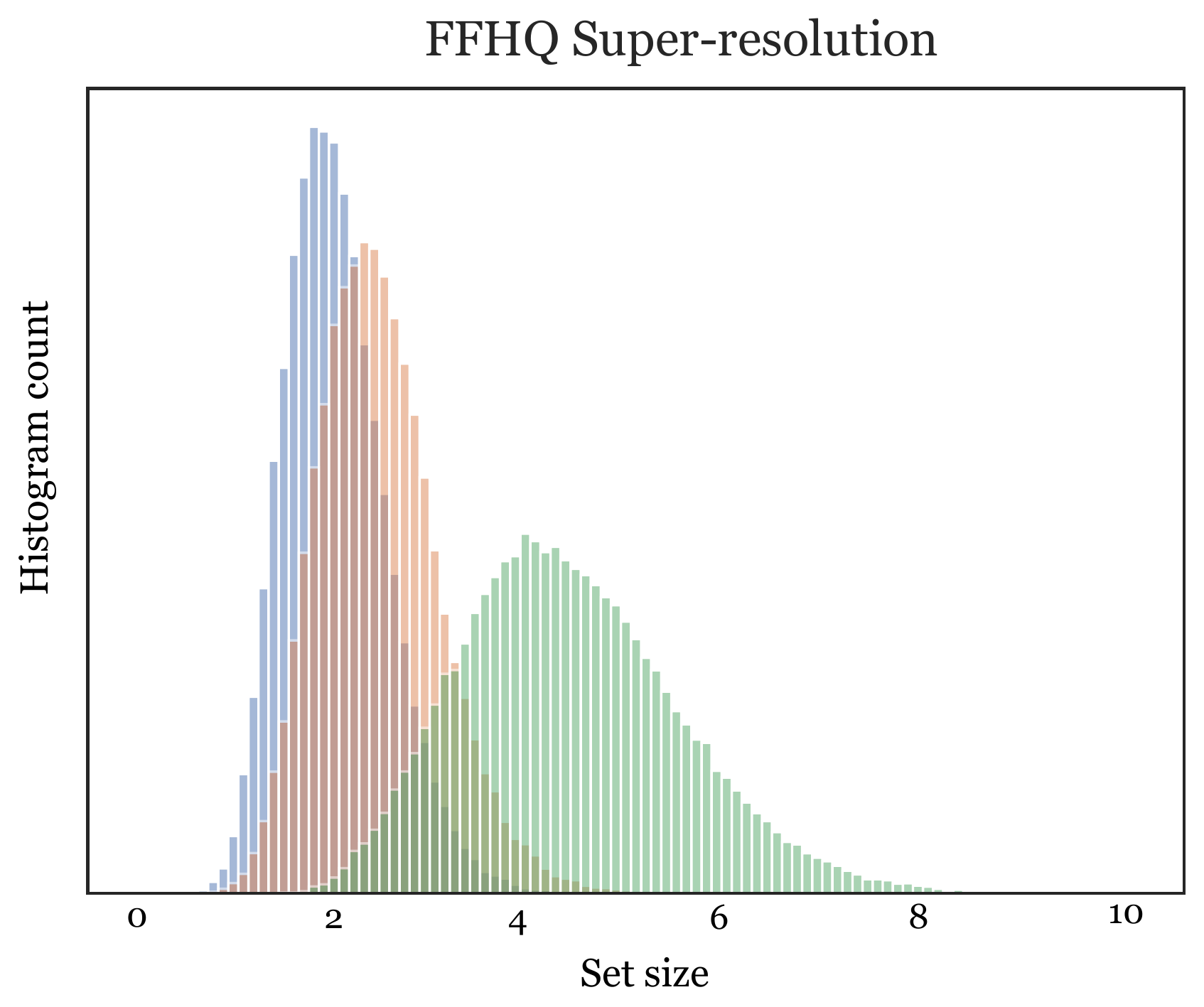} &  
  \includegraphics[width=0.45\textwidth]{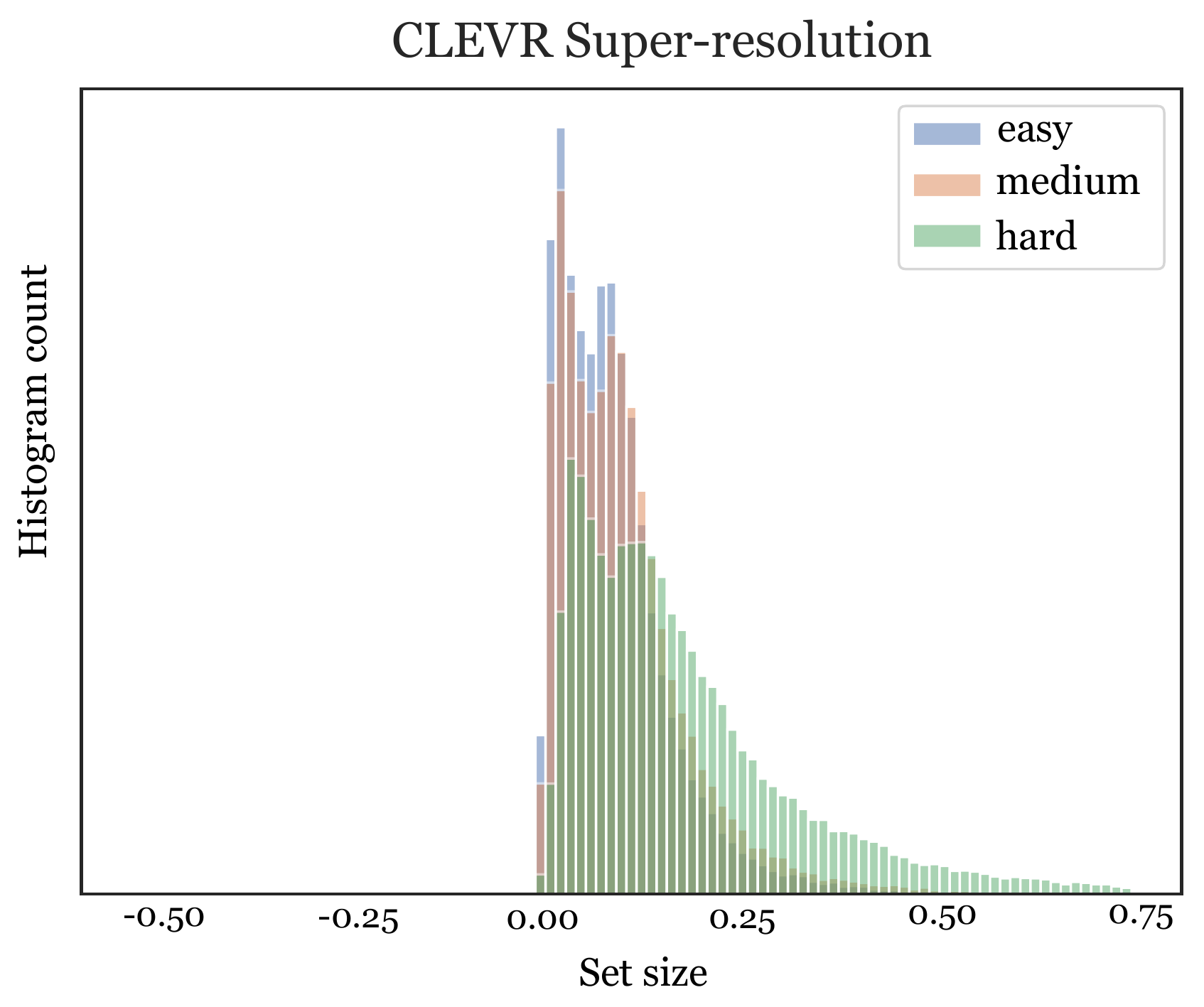} 
\end{tabular}
\caption{\textbf{Adapting to varying corruption levels:}  Distribution of set-sizes for different input corruption levels for super-resolution on FFHQ and CLEVR.}
\label{fig:set_sizes}
\end{center}
\end{figure}
\vspace{-3mm}

\textbf{Description of experiment.}
To simulate varying levels of corruption for image super-resolution, we create the difficulty levels $\{\mathrm{easy},\mathrm{ medium }, \mathrm{hard}\}$ that correspond to $\{1\text{x}, 8\text{x}, 32\text{x}\}$ downsampled versions. All downsampled versions are resized to the same dimensions before presenting to the encoder. The results are computed on a set of 2000 images sampled from the validation split of the CelebA-HQ dataset. In order to obtain the set size for each input, we scale the quantile width using the threshold obtained after the RCPS calibration procedure.

\textbf{Results.}
Figure~\ref{fig:set_sizes} shows the set sizes for the super-resolution corruption model as a function of problem difficulty on two datasets, FFHQ and CLEVR. As expected, the set sizes increase with increasing problem difficulty indicating increasing uncertainty as corruption level increases.

\section{Related Work}
\label{sec:related_work}
\vspace{-0.2cm}
\textbf{GANs for Inverse problems.} 
The remarkable image generation properties of recent approaches such as BigGAN~\cite{bigGAN2019} and StyleGAN~\cite{karras2020stylegan2} has led to the increasing use of these models to solve inverse problems relating to image restoration such as image super-resolution and completion. All prior methods that use GANs for inverse problems such as the ones that use the pretrained generative model as an image prior~\cite{imageprior2020,semanticimageprior2020,pulse2020,brgm2020} or the ones that train an encoder model to project the input into the generator's latent space~\cite{indomain_editing2020,elad2021psp,e2e2021,ohayon2021high} focus on the accuracy of the point estimate but not on the uncertainty level of the input. 

\textbf{GANs for Interpretability.}
Despite providing no guarantees of image likelihood, unlike others such as Normalized Flow~\cite{glow2018} or Score-based models~\cite{scoremodels2021}, GANs have been used to develop interpretable approaches to image generation. The widespread use of GANs as opposed to other generative models in the interpretability is done due to the availability of an disentangled latent space~\cite{gancontrols2020,stylespace2021}, which is a property we utilize in our work.



\textbf{Quantile Regression.} Quantile regression was first proposed in~\cite{koenker1978regression}.
Since then, many papers have used the technique, applying it to machine learning~\cite{hwang2005simple,meinshausen2006quantile,natekin2013gradient}, medical research~\cite{armitage2008statistical}, and more.
Most relevant to us is conformalized quantile regression~\cite{romano2019conformalized}, which gives quantile regression a marginal coverage guarantee using conformal prediction.
Our work instead uses risk-controlling prediction sets, a different distribution-free uncertainty quantification technique.





\textbf{Conformal prediction and distribution-free uncertainty quantification.}
At the core of our proposed method is the distribution-free, marginal risk-control technique studied in~\cite{bates2021distribution} and~\cite{angelopoulos2021learn}.
These ideas have their roots in the distribution-free marginal guarantees of conformal prediction, proposed in~\cite{vovk1999machine}.
Conformal prediction is a flexible method for forming prediction sets that satisfy a marginal coverage guarantee, under no assumptions besides the exchangeability of the test point with the calibration data~\cite{vovk1999machine,vovk2005algorithmic,lei2013conformal,lei2013distribution,lei2014classification,angelopoulos2021gentle}. 
Conformal prediction has been studied in computer vision~\cite{hechtlinger2018cautious,cauchois2020knowing,angelopoulos2020sets,romano2020classification,angelopoulos2021private}, natural language~\cite{fisch2020efficient}, drug discovery~\cite{fisch2021few}, criminal justice~\cite{berk2020improving}, and more.

We are not aware of work applying the notions of conformal prediction and quantile regression to the latent space in generative models. 
\vspace{-0.3cm}
\section{Conclusion}\label{sec:conclusion}
\vspace{-0.3cm}
Experiments indicate that with an appropriately disentangled model, latent space uncertainty intervals express a useful semantic notion of uncertainty previously unavailable in computer vision.
Limitations of our method include that the calibration data must be reflective of the data distribution, and that we assume access to a disentangled latent space.
We see our work as part of a larger tapestry of results in generative models, and our technique will remain applicable as progress gets made on disentanglement and backprojection. Furthermore, by introducing a notion of statistical rigor to GAN based approaches, our work creates a path for application of such models to safety critical settings.
\vspace{-0.3cm}
\section{Ethics}\label{sec:ethics}
\vspace{-0.3cm}
The ethics of generative modeling itself has been called into question given recent events, \textit{e.g.}, the development of deep fakes.
Nonetheless, we believe the downstream consequences of this work will likely be positive.
The techniques herein do not change the predictions of a generative model; they simply provide a calibrated notion of its uncertainty in a relevant semantic space.
Thus, the standard criticism of generative modeling---that it will enable widespread deep fakes---is not applicable.
Furthermore, we expect having a statistically valid and semantically rich notion of uncertainty will provide a sobering reliability assessment of these models, perhaps mitigating the chance of harmful failures.
Finally, although we use face datasets due to their ubiquity in this literature, we have attempted to ethically treat topics like gender and race where they arise.

\subsection*{Acknowledgements}

 A.~N.~A.~was supported by NSF GRFP. S.~B.~was supported by the Foundations of Data Science Institute and the Simons Institute. Y.~R.~was supported by the Israel Science Foundation (grant No. 729/21). Y.~R.~thanks the Career Advancement Fellowship, Technion, for providing research support. S.~S.~ and P.~I.~'s research for this project was sponsored by the United States Air Force Research Laboratory and the United States Air Force Artificial Intelligence Accelerator and was accomplished under Cooperative Agreement Number FA8750-19-2- 1000. The views and conclusions contained in this document are those of the authors and should not be interpreted as representing the official policies, either expressed or implied, of the United States Air Force or the U.S. Government. The U.S. Government is authorized to reproduce and distribute reprints for Government purposes notwithstanding any copyright notation herein. S.~S.~ also acknowledges the MIT SuperCloud and Lincoln Laboratory Supercomputing Center for providing compute resources that have contributed to the results reported in this work.

\bibliography{references}
\bibliographystyle{plain}

\clearpage
\section*{Checklist}


\begin{enumerate}

\item For all authors...
\begin{enumerate}
  \item Do the main claims made in the abstract and introduction accurately reflect the paper's contributions and scope?
    \answerYes{}
  \item Did you describe the limitations of your work?
    \answerYes{We explicitly state the assumptions in Section~\ref{sec:conclusion}}
  \item Did you discuss any potential negative societal impacts of your work?
    \answerYes{Explicitly address this in Section~\ref{sec:ethics}}
  \item Have you read the ethics review guidelines and ensured that your paper conforms to them?
    \answerYes{Explicitly address this in Section~\ref{sec:ethics}}
\end{enumerate}

\item If you are including theoretical results...
\begin{enumerate}
  \item Did you state the full set of assumptions of all theoretical results?
    \answerYes{The proposition is stated in Section~\ref{subsec:calibration}. We also state the assumptions in Section~\ref{sec:conclusion}. For more details, we redirect the reader to previous work where we derive the framework from. (~\cite{bates2021distribution} and~\cite{angelopoulos2021learn}})
        \item Did you include complete proofs of all theoretical results?
    \answerNo{The proposition is stated in Section~\ref{subsec:calibration}. For complete proofs, we redirect the reader to previous work where we derive the framework from. (~\cite{bates2021distribution} and~\cite{angelopoulos2021learn})}
\end{enumerate}

\item If you ran experiments...
\begin{enumerate}
  \item Did you include the code, data, and instructions needed to reproduce the main experimental results (either in the supplemental material or as a URL)?
    \answerYes{Training details are provided in Section~\ref{sec:experiments}. More complete details about the model architecture are provided in the supplementary material. The datasets we use are publicly available, we include references to that in Section~\ref{sec:experiments}. The code will be released in the near future. }
  \item Did you specify all the training details (e.g., data splits, hyperparameters, how they were chosen)?
    \answerYes{}
        \item Did you report error bars (e.g., with respect to the random seed after running experiments multiple times)?
    \answerNo{While we expect there to be some randomness, the behavior of the encoder is consistent across different runs.}
        \item Did you include the total amount of compute and the type of resources used (e.g., type of GPUs, internal cluster, or cloud provider)?
    \answerNo{}
\end{enumerate}

\item If you are using existing assets (e.g., code, data, models) or curating/releasing new assets...
\begin{enumerate}
  \item If your work uses existing assets, did you cite the creators?
    \answerYes{}
  \item Did you mention the license of the assets?
    \answerNo{We cite the dataset creators who have included their respective licenses along with the dataset. We ensure that we follow the terms of the license i.e. use the data for research use only.}
  \item Did you include any new assets either in the supplemental material or as a URL?
    \answerNo{}
  \item Did you discuss whether and how consent was obtained from people whose data you're using/curating?
    \answerNo{Since we use a dataset that comes with a reseach license, we defer such concerns to the dataset creators.}
  \item Did you discuss whether the data you are using/curating contains personally identifiable information or offensive content?
    \answerNo{Since we use a dataset that comes with a reseach license, we defer such concerns to the dataset creators.}
\end{enumerate}

\item If you used crowdsourcing or conducted research with human subjects...
\begin{enumerate}
  \item Did you include the full text of instructions given to participants and screenshots, if applicable?
    \answerNA{}
  \item Did you describe any potential participant risks, with links to Institutional Review Board (IRB) approvals, if applicable?
    \answerNA{}
  \item Did you include the estimated hourly wage paid to participants and the total amount spent on participant compensation?
    \answerNA{}
\end{enumerate}

\end{enumerate}

\appendix

\section{Model architectures}

\subsection{Face experiments}
For the encoder, we use a resnet-50 backbone followed by projection heads that output pointwise, lower and upper quantile predictions. Each projection head consists of a convolution layer followed by a Leaky-Relu activation and a global average pooling layer. The input to each projection head is the output of the backbone network -- a feature map of size $512 \times 4 \times 4$ and the output dimension is the number of style dimensions -- in the case of the pretrained FFHQ styleGAN2 used in our experiments, this value is 9088. 

For the generator, we use a FFHQ pretrained styleGAN2 trained to output faces of resolution $1024 \times 1024$ obtained from the official implementation. No discriminator is used during training.

\subsection{CLEVR experiments}
For the encoder, we use a resnet-18 backbone followed by projection heads that output pointwise, lower and upper quantile predictions. Each projection head consists of a convolution layer followed by a Leaky-Relu activation and a global average pooling layer. The input to each projection head is the output of the backbone network -- a feature vector of size $512$ and the output dimension is the number of style dimensions -- in the case of the pretrained CLEVR styleGAN2 used in our experiments, this value is 204. 

For the generator, we use a modified version of styleGAN2 that is trained to output images of resolution $128 \times 128$. In order to have a controlled latent space, we reduce the size of the style vectors from $512$ in the original model to $12$. This was done to reduce the size of the resulting style dimension from $9088$ to $204$. Since the model was trained on the CLEVR dataset which has less variability compared to other datasets such as FFHQ, the model was able to converge successfully even at this reduced capacity.

\section{Training details}

\subsection{Input preprocessing}
For the face experiments, the inputs to the encoder are resized to $256 \times 256$ and are rescaled to $[-1, 1]$ range. For the super-resolution experiment, the original input is first downsampled as required (i.e. 8x/16x etc) and is resized back to the input resolution $256 \times 256$. For the image inpainting experiment, the corruption mask is generated using the procedure outlined in Section~\ref{subsec:mask_generation}. The image input is then masked to only expose the unmasked parts -- hence the corruption; the mask is concatenated along with the image as an additional input. Example of a masked image is shown the main manuscript in Figure 4.

The procedure described above is repeated for the CLEVR epxeriments with the exception that the input size is $128 \times 128$.

\subsection{Mask generation procedure for image inpainting}\label{subsec:mask_generation}
For generating a corruption model for image completion, we generate a binary mask in a controlled manner. For each input image of size $H \times W \times C$, we start by generating a random mask of size $H \times W$ where each pixel value in contained in the interval $[0, 1]$. For each difficulty level as mentioned in the manuscript (\textit{easy, medium, hard}), we activate only those pixels in the mask whose values lie less than a corresponding threshold. For eg: for the \textit{easy} level, we mask the pixels whose values are less than 0.3. By changing this threshold, we can vary the difficulty level of the masked input. We use the following thresholds: $\{easy: 0.3, medium: 0.6, hard: 0.9\}$. These thresholds were obtained by visual inspection. Intuitively, the threshold can be interpreted as the fraction of pixels that are masked at a given difficulty level -- 30\% being the easier case and 90\% being the harder case. 

\begin{figure}[!h]
\vskip 0.2in
\begin{center}
\begin{tabular}{cc}
  \includegraphics[width=0.7\textwidth]{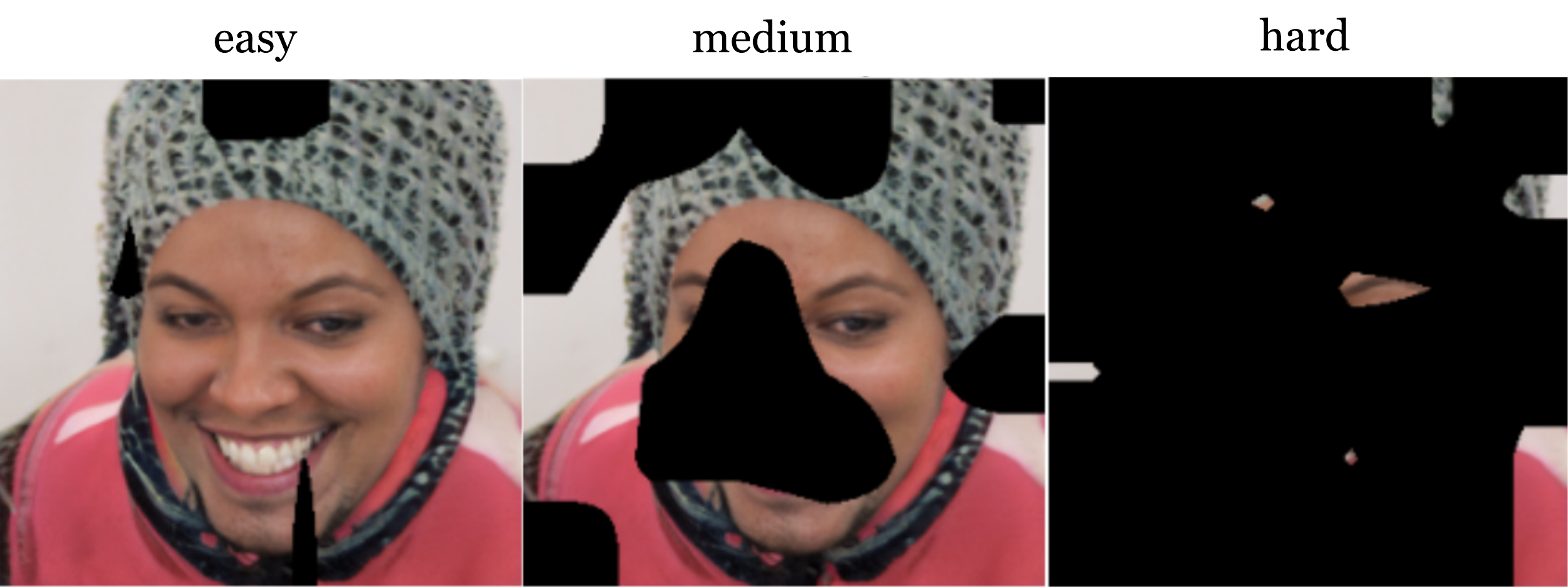} 
\end{tabular}
\caption{\textbf{Inpainting masks:} Masked inputs at different difficulty levels.}
\label{fig:inpainting_masks}
\end{center}
\end{figure}

\subsection{Masking irrelevant style dimensions} \label{subsec:style_masking}
In StyleGAN models, the style space vector is very high dimensional. However, previous work on style space analysis [39] has shown that only few of those dimensions are reliably disentangled. In order to better focus the encoder's capacity only on the disentangled dimensions, we mask out the irrelevant dimensions ensuring that the quantile loss is only applied to the disentangled dimensions. 

For instance, for a FFHQ pretrained model trained to produce an output of size $1024\times1024$ has a style space dimension of 9088.  In order to better focus the encoder's capacity only on the disentangled dimensions, we mask out the irrelevant dimensions. More concretely, we apply the loss function described in ~\eqref{eq:quantile-loss} to the masked latent, $\mathcal{L}_{q_{\beta}}(m \odot x, m \odot z)$, with $m$ being the mask that contains `1' for the disentangled dimensions and `0' otherwise and $\odot$ indicating element-wise product. Note that the masking is applied only to the quantile loss and not the pointwise loss in ~\eqref{eq:l1-loss}. This ensures that the pointwise prediction is able to match the true latents accurately, while the quantile heads focus on learning variablity only in the disentangled dimensions.

\section{Quantitative analysis of interval variability}

In this experiment, we set out to measure the variability of the predicted quantile intervals as a function of problem difficulty. For this analysis, we use 500 images at each difficulty level sampled from the FFHQ-trained pretrained GAN as inputs to our encoder.

For each input, we compute the calibrated uncertainty interval using our approach and compute the \emph{identity} loss specified in Equation~\ref{eq:id-loss}, and \emph{perceptual} loss between the upper and lower edges. We repeat the procedure for each image by varying the input difficulty, similar to the previous Appendix. It can be observed from Table~\ref{tab:metrics} that both perceptual and ID losses increase with increasing perceived input difficulty. This substantiates our claim that the calibrated quantiles display more variability as we increase the difficulty of the task. Note that most of the style dimensions only affect attributes like hair color/glasses/facial hair that do not necessarily change the identity of the individual. Given this observation, the change in ID loss is very much indicative of the variability of the quantile predictions.

\begin{table}[ht]
\caption{\textbf{Measuring variability over quantiles:} Perceptual loss (L-PIPS) and ID Loss between the upper and lower calibrated quantiles.}
\begin{center}
\begin{sc}
\begin{small}
\begin{tabular}{lccc}
\hline
\multicolumn{1}{l}{Metric} & \multicolumn{1}{l}{Easy} & \multicolumn{1}{l}{Medium} & \multicolumn{1}{l}{Hard} \\ \hline
ID Loss              & 0.03                     & 0.06                       & 0.08                     \\ \hline
Perceptual Loss      & 0.17            & 0.21              & 0.24            \\ \hline
\end{tabular}
\label{tab:metrics}
\end{small}
\end{sc}
\end{center}
\vspace{-0.5cm}
\end{table}

\section{Effect of calibration on coverage}

The guarantee in Definition~\ref{def:rcps} tells us that the risk will always be controlled, but it does not tell us that our control will be tight. This experiment tells us how conservative our procedure is, \ie, how closely we match our desired risk and error levels.

\begin{figure}[ht]
\begin{minipage}{0.32\linewidth}
\includegraphics[width=\textwidth]{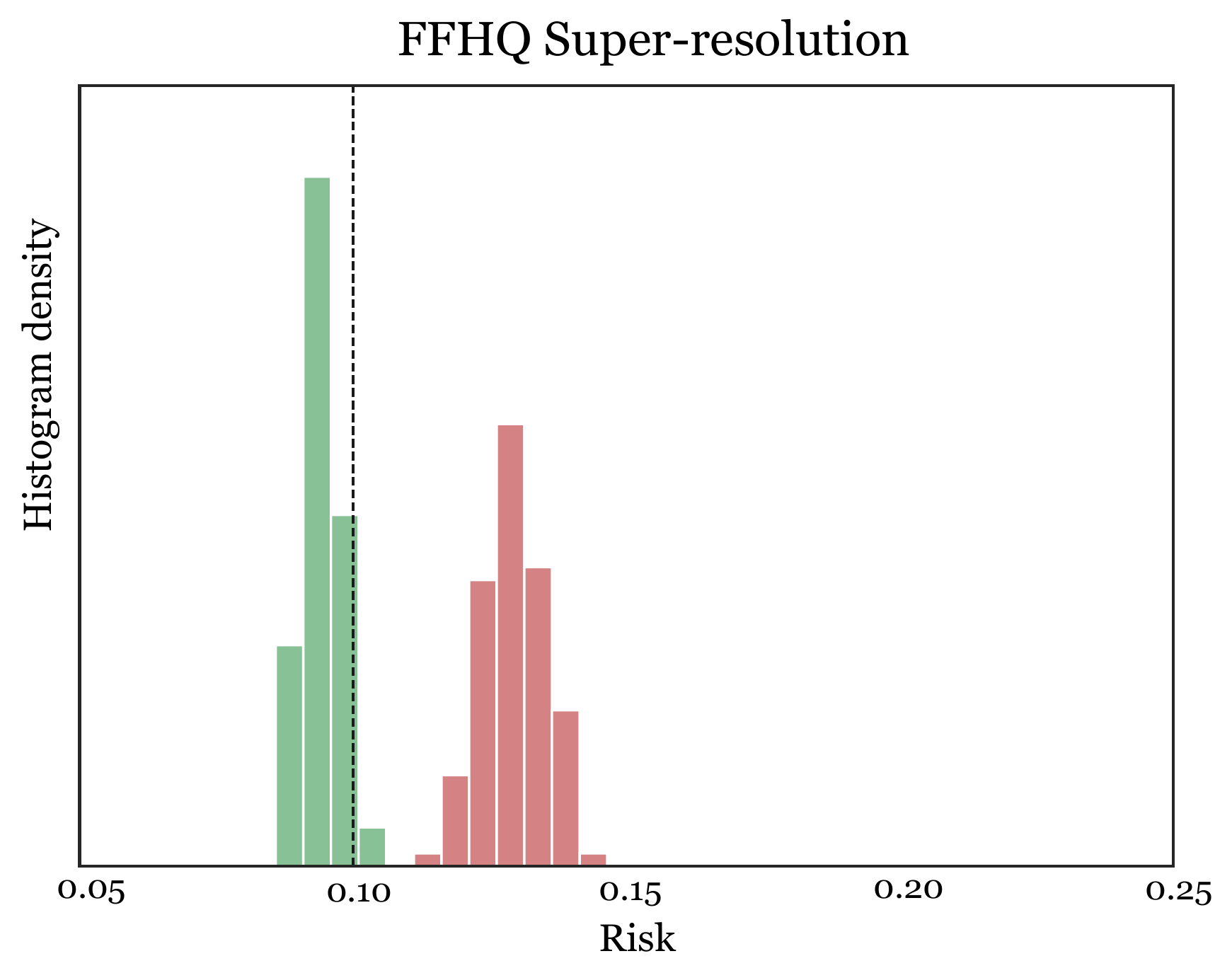}
\label{fig:figure1}
\end{minipage}%
\hfill
\begin{minipage}{0.32\linewidth}
\includegraphics[width=\textwidth]{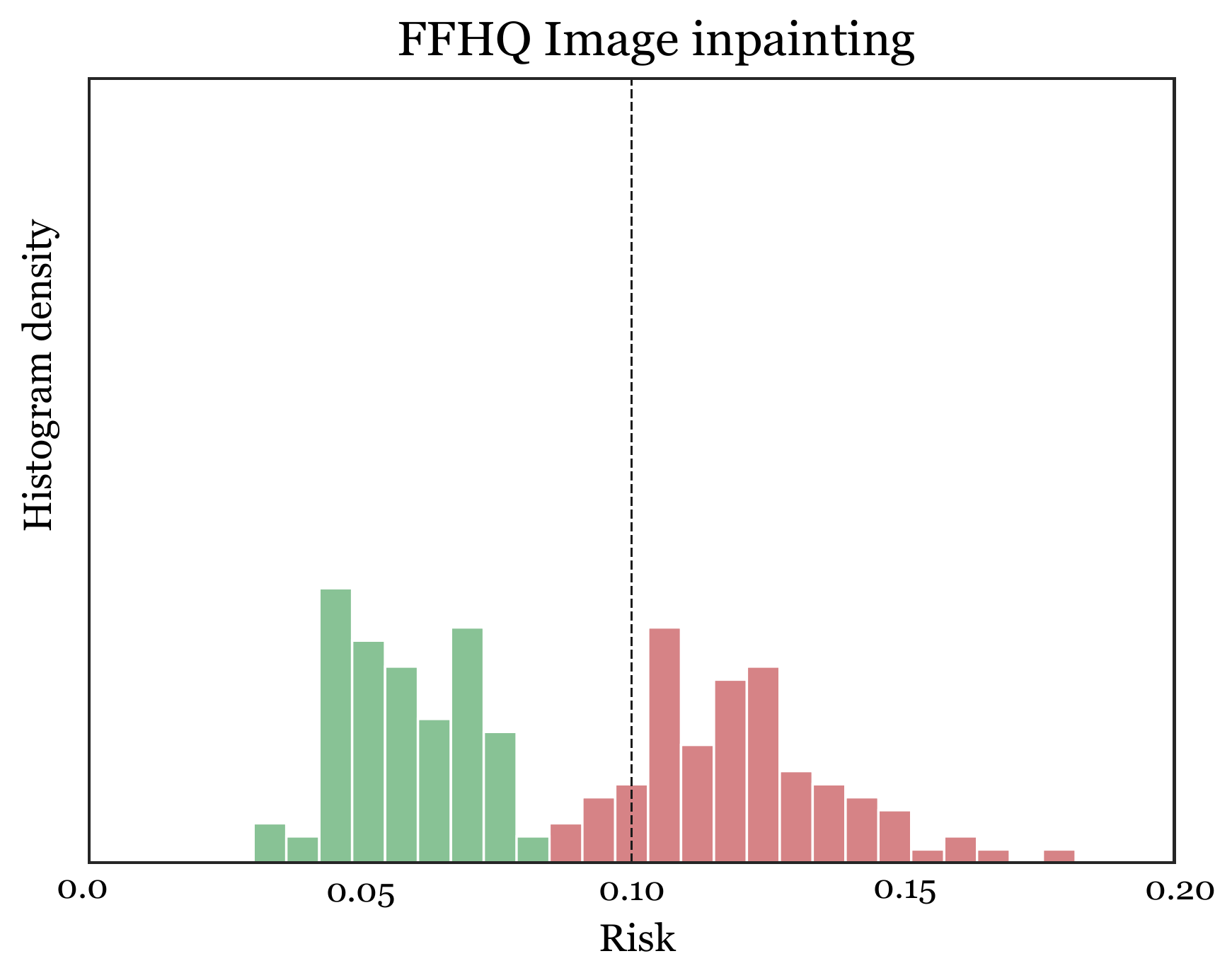}
\label{fig:figure2}
\end{minipage}%
\hfill
\begin{minipage}{0.32\linewidth}
\includegraphics[width=\textwidth]{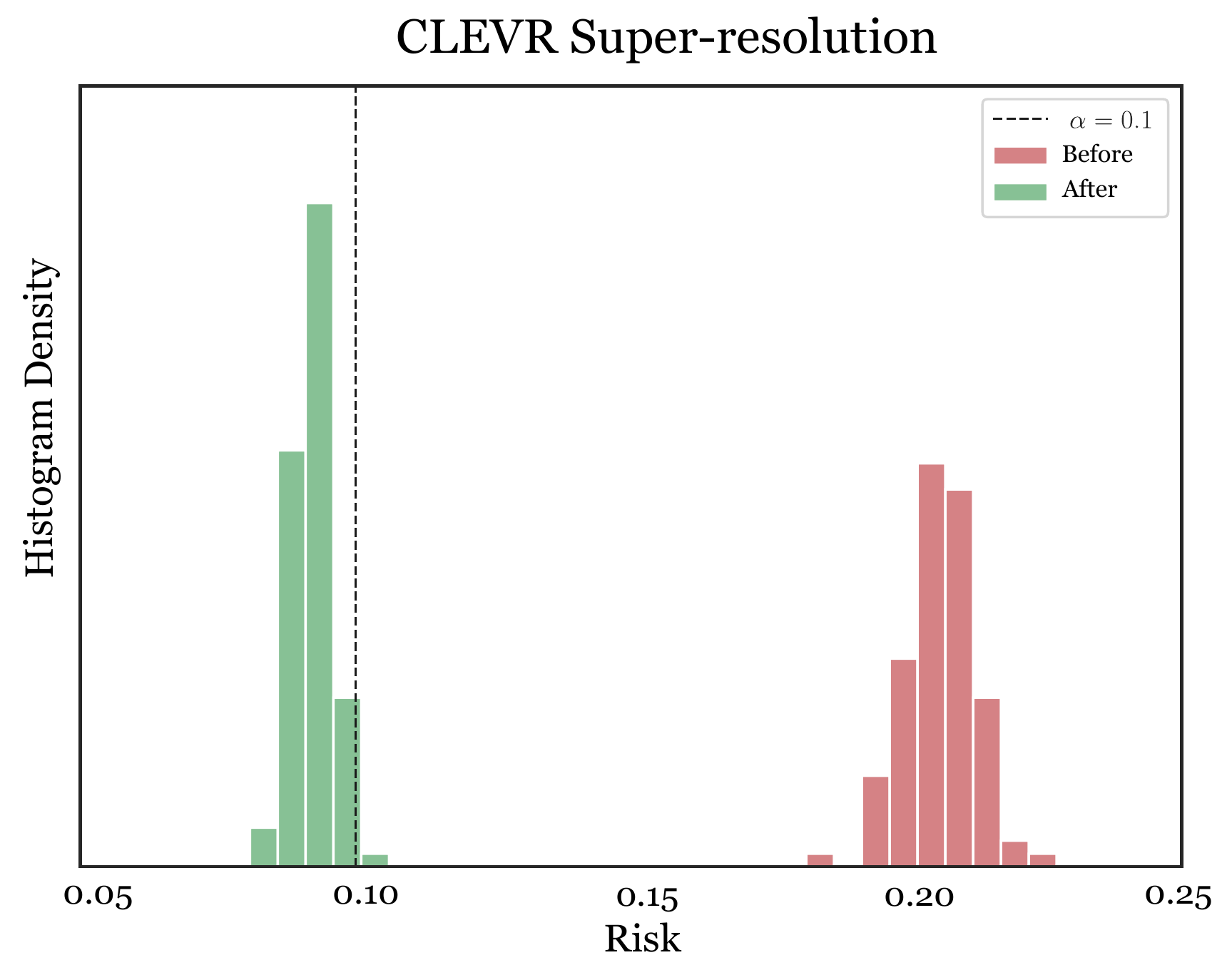}
\label{fig:figure3}
\end{minipage}
\caption{\textbf{Calibration:} Comparison of distribution of empirical risk for 100 calibration runs before and after performing the RCPS calibration procedure. We show results on FFHQ and CLEVR for the Image super-resolution and inpainting corruption models, calibrating for risk level $\alpha=0.1$.}
\label{fig:coverage}
\end{figure}

Since we work in realm of generated data for model training and calibration, we have access to the true latents $Z_d$ which ensures a precise measurement of the average risk. We do a random 50-50 split on the calibration set, where we calibrate on one split and evaluate on the other. To validate the power of the procedure, we repeat this process 100 times. For each run, we report the average risk incurred by our model over the evaluation split. 

Figure~\ref{fig:coverage} compares the average risk of the quantile encoders across different corruption models and datasets, before and after calibration. The performance of the uncalibrated quantile encoder is problem / dataset dependent, \ie, the base model has lower risk in the FFHQ super-resolution problem compared to the inpainting problem or the CLEVR super resolution problem . However, for all settings, the calibration procedure results in lower risk that satisfies the guarantee specified in Definition~\ref{def:rcps}.

\section{Quantifying coverage and adaptivity in real data}

In Table~\ref{tab:adap_cover}, we quantify the effect of coverage and adaptivity on real data in comparison with generated data. The average set size shows that our intervals adapt to problem difficulty successfully for both real and generated data. However the value is higher in case of real data, which is expected since our encoder was not trained for quantile regression on real data. Regarding coverage metrics, we compute the average risk separtely for real and generated data, with and without calibration. While the calibration procedure does not make a significant change in the generated data, since the base encoder is has pretty good coverage to begin with, it makes a significant different in the case of real data.

\begin{table}[H]
\caption{\textbf{Adaptivity and Coverage:} We measure adaptivity by computing the average set size across problem difficulty. Note that the average set size increases with problem difficulty illustrating the adaptivity of the predicted quantile intervals. For quantifying coverage, we measure the average risk before and after calibration. Note that, the risk before calibration on real data is much higher before calibration. This points to the importance of our calibration procedure especially in the presence of real data, which the encoder model was not trained on.}
\label{tab:adap_cover}
\centering
\resizebox{0.7\columnwidth}{!}{%
\begin{tabular}{@{}cccc@{}}
\toprule
\multicolumn{1}{l}{}                       & \multicolumn{1}{l}{} & \multicolumn{1}{l}{\textbf{Generated data}} & \multicolumn{1}{l}{\textbf{Real data}} \\ \midrule
\multirow{2}{*}{\textbf{Average set size}} & Easy                 & 2.5                                       & 3.1                                   \\
                                           & Hard                 & 4.7                                        & 5.3                                   \\ \midrule
\multirow{2}{*}{\textbf{Average risk}}     & w/o calib            & 0.114                                        & 0.267                                   \\
                                           & w/ calib             & \textbf{0.085}                                        & \textbf{0.096}                                   \\ \bottomrule
\end{tabular}%
}
\end{table}

\section{Ablation of feature loss weights}
We ablate the feature loss weights that are specified in Eq~\ref{eq:full-loss}. For simplicity, we fix $c_1=c_2=c$. Table~\ref{tab:ablation} shows the predicted set sizes for different values of $c$. Higher values of $c$ yield more visually pleasing reconstructions during training while also providing slightly tighter quantile sets across varying levels of corruption. Hence, we pick a value of $c=10.0$ for our experiments. A larger hyper-parameter sweep might yield better results.
\begin{table}[ht]
\caption{\textbf{Ablation}: Predicted set sizes for different values of the feature coefficient, \textit{c}. We pick the value $c=10$ as it provides more visually pleasing reconstructions on real data.}
\label{tab:ablation}
\centering
\resizebox{0.4\columnwidth}{!}{%
\begin{tabular}{@{}cccc@{}}
\toprule
\multicolumn{1}{l}{Corruption level} & \multicolumn{1}{l}{c = 0} & \multicolumn{1}{l}{c = 1.0} & \multicolumn{1}{l}{c = 10.0} \\ \midrule
Easy                                 & 2.56                      & 2.4                         & 2.5                          \\ \midrule
Hard                                 & 5.23                      & 5.1                         & 4.7                          \\ \bottomrule
\end{tabular}%
}
\end{table}

\end{document}